\documentclass[letterpaper, 10 pt, conference]{ieeeconf}  

\IEEEoverridecommandlockouts                              

\usepackage{amsmath} 
\usepackage{amssymb}  
\usepackage[linesnumbered,ruled]{algorithm2e}
\usepackage{gensymb}
\usepackage{graphicx}
\usepackage{epstopdf}
\usepackage{subcaption}
\usepackage[colorlinks]{hyperref}
\usepackage{multirow}
\usepackage[T1]{fontenc}
\usepackage{fix-cm}

\makeatletter
\let\NAT@parse\undefined
\makeatother

\usepackage[square,comma,numbers,sort&compress]{natbib} 

\title{\LARGE \bf
Fast Cylinder and Plane Extraction from Depth Cameras for Visual Odometry
}

\author{Pedro F. Proen\c{c}a$^{1}$ and Yang Gao$^{1}$
\thanks{$^{1}$The authors are with the Surrey Space Centre, Faculty of Engineering and Physical Sciences, University of Surrey, GU2 7XH Guildford,
	U.K. {\tt\small \{p.f.proenca, yang.gao\}@surrey.ac.uk}}%
}

\begin{document}

\maketitle
\thispagestyle{empty}
\pagestyle{empty}

\begin{abstract}
This paper presents CAPE, a method to extract planes and cylinder segments from organized point clouds, which processes 640$\times$480 depth images on a single CPU core at an average of 300 Hz, by operating on a grid of planar cells. While, compared to state-of-the-art plane extraction, the latency of CAPE is more consistent and 4-10 times faster, depending on the scene, we also demonstrate empirically that applying CAPE to visual odometry can improve trajectory estimation on scenes made of cylindrical surfaces (e.g. tunnels), whereas using a plane extraction approach that is not curve-aware deteriorates performance on these scenes. \par
To use these geometric primitives in visual odometry, we propose extending a probabilistic RGB-D odometry framework based on points, lines and planes to cylinder primitives. Following this framework, CAPE runs on fused depth maps and the parameters of cylinders are modelled probabilistically to account for uncertainty and weight accordingly the pose optimization residuals.
\end{abstract}



\section{Introduction}

\begin{figure}[t]
\centering
	\begin{tabular}{@{}c@{ }c@{}}
		\includegraphics[scale=0.4]{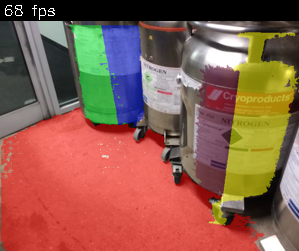} &
		\includegraphics[scale=0.4]{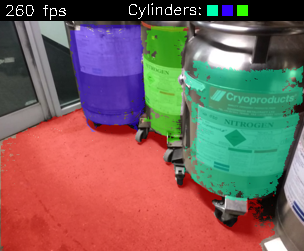}\\
		\includegraphics[scale=0.4]{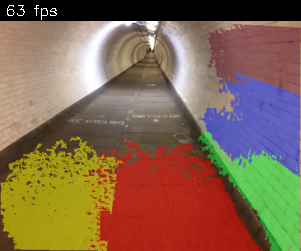} &
		\includegraphics[scale=0.4]{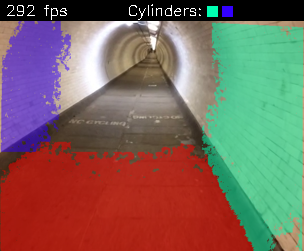}\\
		PEAC \cite{feng2014fast} & CAPE (this work)
	\end{tabular} 
	\caption{Ouput of our method and state-of-the-art plane extraction method PEAC on cylindrical surfaces. While our method can capture adequately these primitives, the planes fitted on these surfaces by PEAC are unstable and can hence degrade camera pose estimation as shown in this paper. Both methods were used with a patch size of 20 $\times$ 20 pixels. A video of sequences processed by these methods is available at: \url{https://youtu.be/FPFPVwm_yq0}.}
	\vspace*{-1mm} 
	\label{fig1}
\end{figure}

Man-made environments are predominantly made of planar surfaces, thus a recent trend in SLAM \cite{PlaneAndsurfels_2014,Taguchi2013,kpaslam,ma2016cpa,proencca2017probabilistic} for structured environments is to exploit plane primitives to reduce drift, reconstruct compact maps, and improve the robustness of visual odometry (VO). Such systems that are based on RGB-D cameras start typically by extracting plane segments from the depth map, by employing plane segmentation algorithms such as the one shown in Fig. \ref{fig1}.
This method \cite{feng2014fast} achieves good accuracy in most environments while being significantly faster than other alternatives \cite{holz2013fast,trevor2013efficient}, which is a requirement for real-time applications. However, it fits incorrectly plane segments to smooth curved surfaces by using a greedy clustering algorithm. Such plane segments are unstable and can thus deteriorate the performance of camera pose estimation. This issue is more severe in industrial and underground environments as these tend to be made of cylindrical surfaces.

To address these environments, we first propose a cylinder\footnote{Cylinder is defined as an infinite surface instead of a solid.} and plane extraction method, named CAPE, which operates efficiently on a grid of planar cells by performing cell-wise region growing guided by a histogram of cell normals, and then using a model fitting scheme that classifies the shape of segments based on principal component analysis (PCA) and uses a direct solution based on linear least squares to cylinder fitting, embedded in sequential RANSAC. Moreover, the boundary of segments is refined approximately by again exploiting the cell grid. We believe that in this kind of applications, guaranteeing low latency is more important than obtaining precisely the exact segment boundaries.

Secondly, we propose to use the cylinders primitives, given by CAPE, along with other features, for camera pose estimation by extending a probabilistic RGB-D Odometry framework  \cite{proencca2017probabilistic} that already uses points, lines and planes. Following this framework, cylinder parameters are modelled probabilistically to account for uncertainty and pose is estimated by aligning cylinder axes.
Experiments were carried out both on RGB-D sequences captured in environments with cylinder surfaces (e.g. tunnel) and without them. Our results show that applying the cylinders, extracted by CAPE, to VO improves performance on scenes made of cylindrical surfaces whereas using just the planes given by PEAC \cite{feng2014fast} deteriorates the performance of baseline on these scenes. Furthermore, CAPE is on average 4-10 times faster than PEAC, depending on the scene and has a more consistent latency, around 3 ms. The source code of CAPE is available at: \url{https://github.com/pedropro/CAPE}.

\begin{figure*}
\centering
		\includegraphics[scale=0.348]{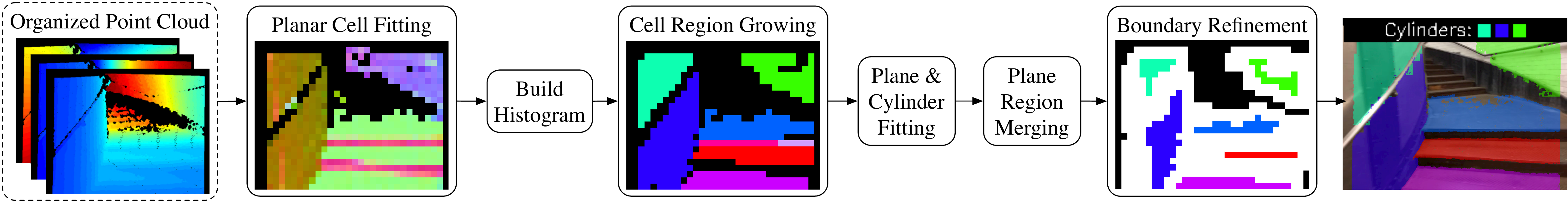}
	\caption{Overview of CAPE main processes. Normals of planar cells are color-coded on the second image. On the last image the refined segments are overlaid on the respective RGB image.}
	\label{fig2}
	\vspace*{-1mm} 
\end{figure*}

\section{Related Work}

Three techniques are often used in shape primitive extraction from point clouds: 
RANSAC, Hough transform and Region Growing. RANSAC has been widely used for plane extraction \cite{yang2010plane,Taguchi2013,biswas2012depth} and it was further used in \cite{schnabel2007efficient} for extracting spheres, cylinders, cones and tori. However, the RANSAC algorithm does not exploit spatial (i.e. connectivity) information thus these approaches enforce locality by constraining the sampling area. Hough transform, originally proposed for 2D line detection, was applied to plane and cylinder extraction in \cite{vosselman2004recognising,rabbani2005efficient} by using a Gauss map. A more efficient Hough transform voting scheme was proposed, in \cite{limberger2015real}, whereby votes are cast by planar clusters given by an octree.

In contrast to these approaches, region growing exploits the connectivity information. For example, in \cite{poppinga2008fast}, plane segments are grown point-wise by recursively adding at each iteration a neighbouring point, fitting a new plane  and checking if the respective mean square error (MSE) is low enough to accept the point. Effort was made to implement this efficiently, e.g., using a priority search queue, however the merging attempts are still costly for the amount needed as they involve an eigen decomposition for plane fitting besides the nearest neighbour search. To reduce this merging attempt cost, \cite{holz2013fast} instead suggests updating the plane parameters approximately by just averaging the point normals. This requires point normal computation, which is known to be costly, however, for organized point clouds, \cite{holz2011real} proposed an efficient solution by exploiting the image structure. This was used in \cite{trevor2013efficient} with a connected component labelling for plane segmentation and to achieve real-time performance (30 Hz) the normal estimation and plane segmentation are processed in parallel.
The computational cost of these methods can be significantly reduced by operating on point (pixel) clusters instead of point-wise operations:
A real-time solution is proposed in \cite{feng2014fast}, referred to as PEAC, which achieves state-of-the art segmentation results by operating on a graph of image patches. Plane segments formed of patches are grown using a hierarchical agglomerative clustering method and then these are refined using pixel-wise region growing. Our method follows this approach by operating on a grid of planar cells, but achieves higher efficiency by avoiding pixel-wise region growing and the clustering algorithm. A limitation of this clustering approach and \cite{poppinga2008fast} is that planes are greedily fit to any smooth curved surfaces, whereas for example, in \cite{trevor2013efficient}, segments are discarded by analyzing the segment covariance.\par
Several SLAM systems using plane primitives have emerged recently \cite{Taguchi2013,PlaneAndsurfels_2014,ma2016cpa,kpaslam,proencca2017probabilistic}. While, \cite{PlaneAndsurfels_2014} showed how to compress dense reconstructed models by using plane segments, drift is reduced in \cite{ma2016cpa,kpaslam} by using plane landmarks in a map optimization framework. Recently, we also demonstrated in \cite{proencca2017probabilistic} that combining planes with other geometric entities (i.e. points and lines) improves the robustness of VO particularly to low-textured surfaces. That work  \cite{proencca2017probabilistic} introduced a feature-based RGB-D Odometry framework that uses probabilistic model fitting and depth fusion to derive and model uncertainty for frame-to-frame pose estimation. This work extends that probabilistic framework to cylinders. \par Although semantic SLAM methods \cite{salas2013slam++,bowman2017probabilistic} use high-level features (e.g. chairs, doors ) as landmarks in map optimization, exploiting explicitly the geometry of cylinder models for VO or SLAM remains unexplored, with the exception of the fisheye-monocular VO and mapping in \cite{hansen2013pipe} which incorporate pipes as geometric constraints into Sparse Bundle Adjustment.

\section{CAPE: Cylinder and Plane Extraction}
\label{sec:method}

The workflow of our method is illustrated in Fig. \ref{fig2}. Given a point cloud organized in image format, i.e., depth map plus back-projected X and Y coordinates, our method starts by trying to fit planes to pixel patches (grid cells), distributed according to a specified grid resolution. Smooth surfaces are then found by performing region growing on these planar cells, where seeds are selected according to an histogram of normals, built a-priori. Each resulting segment, with enough cells, is then processed by a model fitting algorithm with a cascade scheme for plane and cylinder models. During this process, segments can be split since physically connected primitives can be merged by region growing. On the other hand, plane segments can be merged afterwards if they share similar model parameters and have connected cells. Finally, the boundaries of the segments are refined pixel-wise within cells selected through morphological operations. Details of these modules are given in the following sections.

\subsection{Planar Cell Fitting}
\label{sec:planar_grid}
This step is also performed by the first stage of the graph initialization in \cite{feng2014fast}. Given a uniform grid of non-overlapping patches, we assess the planarity of each patch. First, cells with significant missing points or discontinuous depth are promptly classified as non-planar (seen as black cells in Fig. \ref{fig2}) as planar fitting on these can yield small plane residuals, particularly with small patches. Specifically, we check if the fraction of missing points is more than a certain tolerance, while discontinuity is only checked for depth pixels along a vertical and horizontal line passing through the patch center, such that, if the depth difference between any adjacent valid pixels, within this cross, is more than a certain value, the cell is considered non-planar. Using this approximate search involves less checks than looking at the depth differences of every pixel's neighbourhood as in \cite{feng2014fast}. \par 
Plane fitting is then performed on each cell that passes these conditions by using principal component analysis (PCA) , in which the plane normal is given by the eigenvector with the lowest eigenvalue and the plane's Mean Squared Error (MSE) is given by that eigenvalue. As in \cite{feng2014fast}, the cell is classified as planar if the MSE is less than  $(\sigma_{\bar{z}} +\epsilon)^2$, where $\sigma_{\bar{z}}$ is the estimated standard uncertainty for the cell's mean depth and $\epsilon$ is a tolerance coefficient.
To later, fit planes efficiently on merged cells, cells need to store only the first and second raw moments of 3D points, since the covariance matrix can be conveniently retrieved using the K\"{o}nig-Huygen formula.

\subsection{Histogram of Normals}
\label{sec:histogram}

\begin{figure}
	\centering
	\includegraphics[scale=0.4]{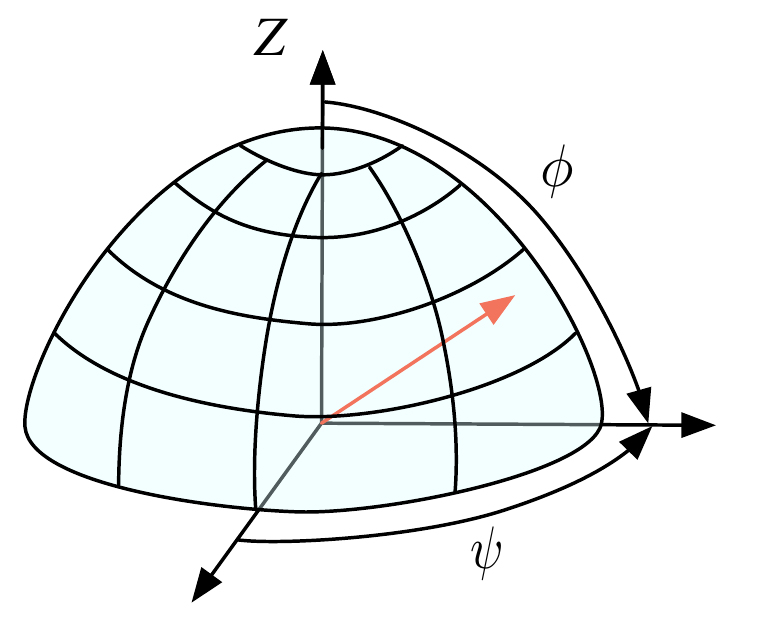}
	\caption{Sphere structure used to build the histogram of cell normals. Polar angle is shown as $\phi$ and azimuth as $\psi$. $Z$ axis points towards the camera plane. Notice how normals with $\phi$ less than one quantization step are assigned the same bin.}
	\label{fig3}
	\vspace*{-1mm} 
\end{figure}

To perform region growing, cell seeds are selected according to the dominant directions of the planar cell normals. To do this, we use a dynamic 2D histogram of normals represented in spherical coordinate angles. Building the histogram involves converting the normal vectors of planar cells to polar and azimuth angles, and quantizing these using the structure shown in Fig. \ref{fig3}. To avoid coordinate singularity, polar angles less than the polar angle quantization step are assigned an azimuth of 0$^\circ$. \par 
During region growing, cells assigned to the most frequent histogram bin are iteratively retrieved and the histogram is updated by removing cells found by region growing.

\subsection{Cell-wise Region Growing}
\label{sec:RG}

The region growing loop is shown in Algorithm \ref{alg1}. The region growing itself implemented by the function $GrowSeed(G, L, s)$ uses 4-neighbour search and proceeds as follows: A cell neighbour $c$ of a current seed $s$ is added to the region $R$ if: (i) it is contained in the list of remaining cells $L$, (ii) the dot product of the cell normals is more than $T_{N}$ and (iii) the point-to-plane distance of the centroid of $c$ to the seed's plane is less than $T_d(s)$, which is pre-computed as: $l \sqrt{(1-T^2_N)}$, where $l$ is the distance between the 3D points at the corners of cell $s$. Then, $T_d(s)$ must be further truncated. This adaptive threshold compensates the fact that the distance between cell centroids increases with the depth and so does the point-to-plane distances if we consider a constant angle between normals of  $\arccos{(T_N)}$.

\begin{algorithm}
    \SetKwInOut{Input}{Input}
    \SetKwInOut{Output}{Output}
    \Input{Grid of planar cells $G$ and histogram $H$}
    \Output{Set of segments $S$}
    $S \longleftarrow \emptyset$; $L \longleftarrow G$\;
    \While{$L \neq \emptyset$}{
    $C \longleftarrow$ GetCellsFromMostFrequentBin($H$)\;
   	\If{$|C|<k_1$}{break\;}
    $s  \longleftarrow $GetCellWithMinMSE($C$)\;
    $R \longleftarrow $  GrowSeed($G$, $L$, $s$)\;
    $L \longleftarrow L \setminus R$\;
    $H  \longleftarrow$ RemoveCellsFromHistogram($H$,$R$)\;
    \If{$|R|<k_2$}{continue\;}
    $S  \longleftarrow$ $S \cup R$\;
    }
    \caption{Cell-wise Region Growing}
    \label{alg1}
\end{algorithm}

\subsection{Plane and Cylinder Fitting}
\label{sec:plane_vs_cylinder}

Our approach to model fitting follows a staged scheme, which is shown in Algorithm \ref{alg2}. First, for each segment, comprised of planar cells, provided by region growing, a plane is fitted by using the raw moments of each cell, as discussed in Section \ref{sec:planar_grid}, to obtain the covariance of all points in the segment. Planarity is assessed by checking the ratio of the second largest eigenvalue to the smallest eigenvalue of this covariance, which is done in line \ref{alg:line3}. If this is large enough, the segment is labelled as a plane and the grid segmentation and its plane parameters are stored. Otherwise, we check if the surface is extruded, i.e., invariant in one direction, which is a property of open cylinders. \par 
Concretely, this can be done by analyzing the distribution of surface normals since in noise-free extruded surfaces, the smallest eigenvalue of the covariance of normals is always zero. Therefore, as shown in line \ref{alg:line6}, given the set of cell normals $N$, we perform PCA on the stacked matrix $[N, -N]$ to compensate the fact that only a fraction of the cylindrical surface is detected. Additionally, the span of this area will affect the second eigenvalue, therefore we choose the ratio of the first $\lambda_{max}$ to the last $\lambda_{min}$ eigenvalues, known as the \textit{condition number}, as a criteria to accept processing a surface. For example, a sphere would fail this test, whereas a segment comprised of several planes or cylinders could pass this test. This is important since oversegmentation may happen in region growing. This is also why a sequential RANSAC algorithm, following the approach of \cite{zuliani2005multiransac}, is used at this stage to fit multiple cylinder models. This implies that cylinders may be fitted to actual planes, which is not an issue since cylinders can be seen as a generalization of planes, in the sense that an infinite plane corresponds to a cylinder with infinite radius. After obtaining one or multiple subsegments of $R$ from the RANSAC process, to find if they belong to either planes or cylinders, a plane is fitted to each subsegment and the respective MSE is compared against the MSE of the cylinder, which is given by point-to-axis distances. In terms of RANSAC implementation, each iteration selects three cells, although two is the minimal case, and uses the solution explained below to find cylinder parameters.

\begin{figure}
	\centering
	\includegraphics[scale=0.6]{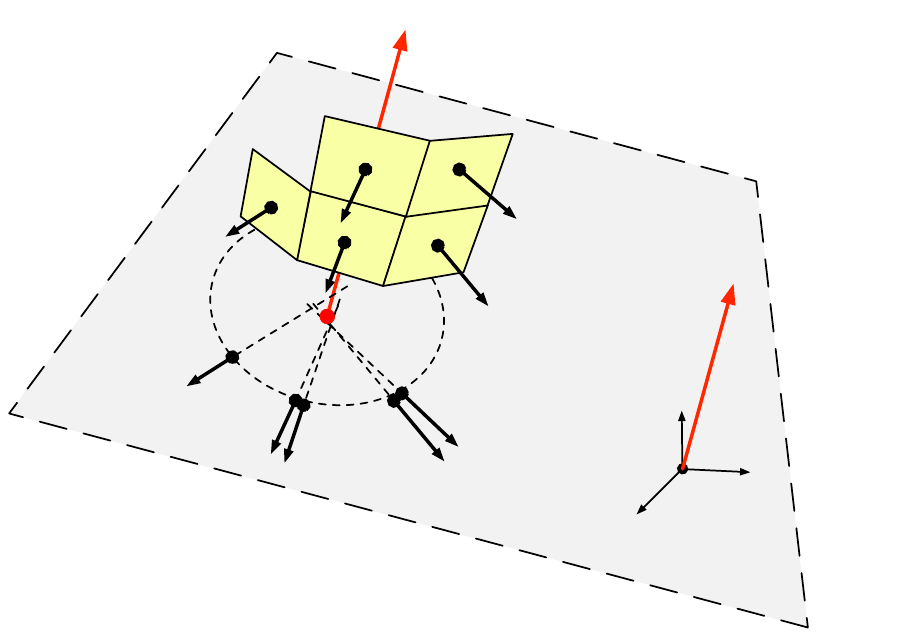}
	\caption{Direct cylinder fitting solution based on cell normals for a convex cylinder. First, cell centroids and normals are projected onto a plane according to the cylinder axis. Second, an analytical circle fitting solution is used to estimate the radius and center.}
	\label{fig4}
	\vspace*{-1mm} 
\end{figure}

Our proposed solution to cylinder fitting based on normals is depicted in Fig. \ref{fig4}. It exploits the fact that surface normals should intersect orthogonally the cylinder axis. Let the column vectors $P_i$ and $N_i$ be respectively the centroid and the plane normal of one cell among the $n$ cells contained in a segment. Then, first, the cylinder axis $v$ is found through the PCA on the normal vectors described above, where $v$ corresponds to the eigenvector with the smallest eigenvalue. To simplify the problem as a circle fitting one, the cell centroids and normals are projected on the plane perpendicular to $v$ passing through the origin of the reference frame. Concretely, a projected point is given by:
\begin{equation}
\label{eq:pl_proj}
P_i' = P_i - v (v \cdot P_i)
\end{equation}
whereas projected normals $N'$ are obtained the same way but then normalized. We can now cast the circle fitting as a 1D linear least squares problem by minimizing:
\begin{equation}
\label{eq:lls_circle}
E = \sum_{i}^{m} \frac{(P'_i-rN'_i-C)^2}{2}
\end{equation}
where $C$ is the circle center and $r$ is the radius. If we set the derivative of (\ref{eq:lls_circle}) wrt. $r$ equal to zero, we arrive at the radius solution:

\begin{equation}
\label{eq:r_sol}
\hat{r} = \Big(1-\frac{1}{m}\sum_{i}^{m} N'^\top_i \bar{N'}\Big)^{-1}\Big(\frac{1}{m} \sum_{i}^{m} N'^\top_i(P'_i-\bar{P'})\Big)
\end{equation}
where $\bar{N'}$ and $\bar{P'}$ represent the respective means. Then, given this estimated radius, the circle center is
\begin{equation}
\label{eq:C_sol}
\hat{C} = \frac{1}{m}\sum_{i}^{m}(P'_i- \hat{r}N'_i)
\end{equation}
This solution is valid only when the normals point outwards. Otherwise, the radius is negative, thus we take its absolute value. 
\par
To apply this method to RANSAC, inliers are detected by using the residual in (\ref{eq:lls_circle}) divided by the estimated radius, since (\ref{eq:lls_circle}) increases with the radius, given noisy normals. Inliers are selected if this relative error is less than 15\% and the MSAC criteria \cite{torr2000mlesac} is used to score each model hypothesis. We have tried alternatively using the point-to-circle distance but we found this was not as discriminative, as it fits a cylinder to several planar surfaces. Finally, the model is refined with all the inliers in (\ref{eq:r_sol}) and (\ref{eq:C_sol}).

\begin{algorithm}
	\SetKwInOut{Input}{Input}
	\SetKwInOut{Output}{Output}
	\Input{Segment R and its cell normals $N$ and centroids $P$}
	\Output{Set of planes $\mathcal{M}$ and set of cylinders $\mathcal{C}$}
	$\mathcal{M} \longleftarrow \emptyset$; $\mathcal{C} \longleftarrow \emptyset$\;
	\textit{plane\_score} $ \longleftarrow$ FitPlane($R$)\;
	\eIf{plane\_score > plane\_min\_score}
	{\label{alg:line3}
		$\mathcal{M}  \longleftarrow \mathcal{M} \cup R$\;
	}
	{
	$\{v, \lambda_{max}, \lambda_{min}\} \longleftarrow $ PCA($-N \cup N$)\;
	\label{alg:line6}
	\If{$\lambda_{max} / \lambda_{min}$ > extrusion\_min\_score}
		{
		$\{P', N'\} \longleftarrow$ ProjectToPlane($P, N, v$)\;
		$\{I\} \longleftarrow$  FitCylinderWithSeqRANSAC($P',N'$)\;
		\ForEach{ \textnormal{subsegment} $I_i \in I$}
		{
		FitPlane($I_i$)\;	
		\eIf{$\textnormal{MSE}_{\textnormal{plane}}(I_i) \leq\textnormal{MSE}_{\textnormal{cylinder}}(I_i)$}
			{$\mathcal{M}  \longleftarrow \mathcal{M} \cup I_i$\;}
			{$\mathcal{C}  \longleftarrow \mathcal{C} \cup I_i$\;}
		}

		}
	}
	\caption{Plane and Cylinder fitting}
	\label{alg2}
\end{algorithm}

\subsection{Model Segment Refinement}
\label{sec:refinement}

The grid segments seen in Fig. \ref{fig2} are quite coarse, thus their boundaries are refined. In PEAC, these are refined by using pixel-wise region growing. Unfortunately, as revealed in the timing results in \cite{feng2014fast}, this step is computationally expensive and moreover it does not guarantee accurate results, as shown in Fig \ref{fig1}, top-left image, as segments can grow unbounded beyond their surface since pixel normals are not considered. Therefore, we propose a cell-bounded search based on morphological operations on the cell grid: First, each grid segment is eroded using a 3$\times$3 kernel (searching element) to remove the boundary cells. This first step is also performed by the refinement algorithm in \cite{feng2014fast}. In this work, we discard segments that are completely eroded, and use a 4-neighbour erosion kernel which is less destructive than the 8-neighbour kernel. Then, the original segment is dilated with an 8-neighbour 3$\times$3 kernel to possibly expand our segment. The cells valid for refinement are given by the difference between the dilated segment and the eroded segment. These are shown as white cells in Fig. \ref{fig2}. The distance between the segment model and each point within these cells is calculated, so that each pixel is assigned to the model if its square distance is less than the model MSE times a constant ($k=9$ in this work) and if it is the minimum distance to any model sharing the refinement cell. Thus, distances need to be stored while the segments are refined.

\section{Cylinders for RGB-D Odometry}

\begin{figure}
	\centering
	\includegraphics[scale=0.4]{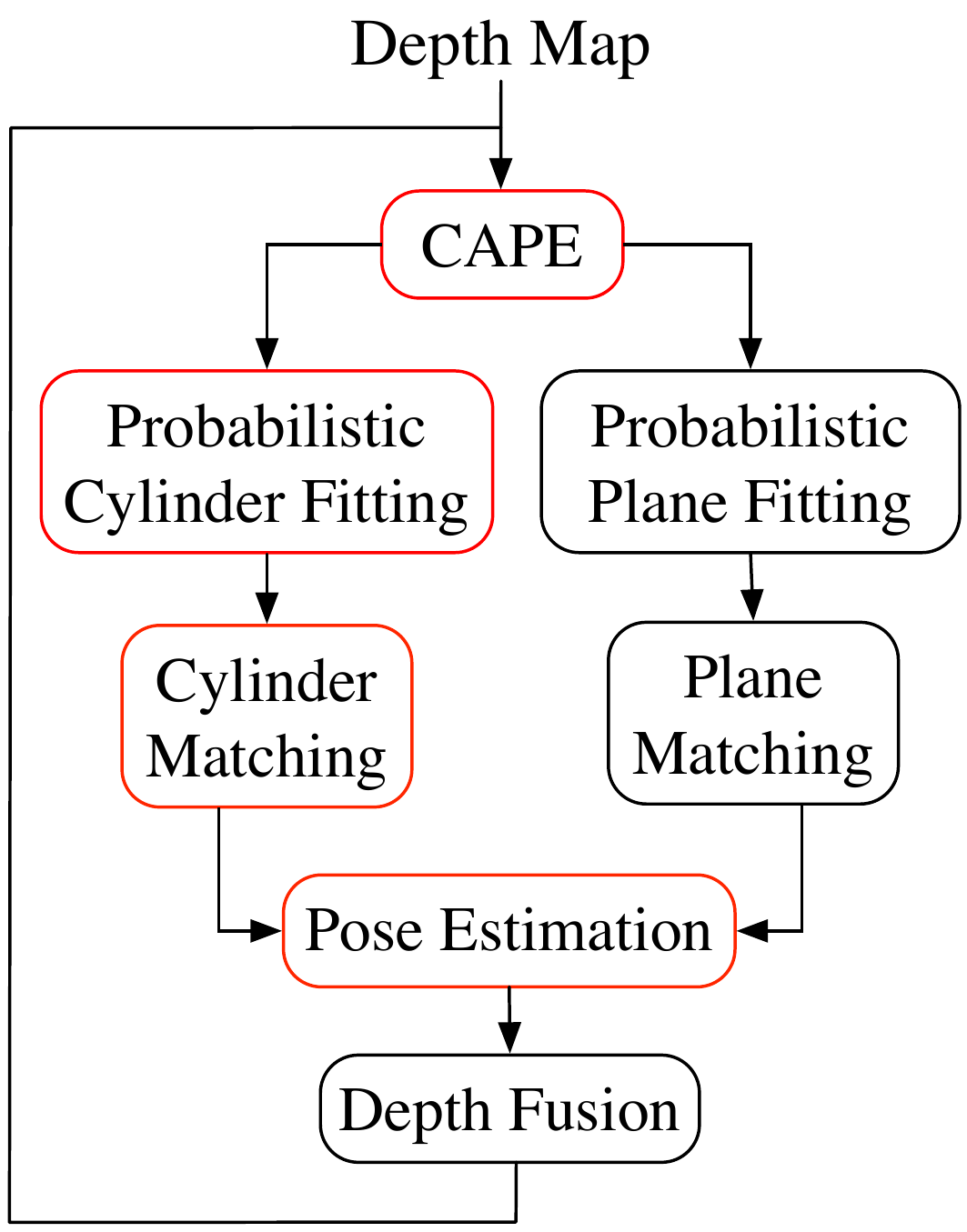}
	\caption{Pipeline of VO in \cite{proencca2017probabilistic} extended by this work. Extensions are colored in red.}
	\label{fig5}
	\vspace*{-1mm} 
\end{figure}

To exploit both the extracted planes and cylinder primitives for VO, we extended the RGB-D odometry framework method in \cite{proencca2017probabilistic} to cylinders. The original system already allows the use of points, lines and plane features within a probabilistic framework that models and propagates the uncertainty of depth and the feature parameters. However, contrary to full SLAM systems and sophisticated VO methods, the pose estimation uses a basic frame-to-frame scheme, which is known to be prone to drift. To address the sensor depth noise, the framework also incorporates a probabilistic depth filter for spatio-temporal depth fusion based on Mixture of Gaussians that models explicitly the depth uncertainty. Fig. \ref{fig5} highlights the extensions made by this paper to this framework , while a more detailed model with points, line segments can be found at \cite{proencca2017probabilistic}. After extracting plane and cylinder segments and their parameters, probabilistic model fitting solutions are used to refine the parameters and derive their uncertainty, then cylinders and planes are matched heuristically between last and current frame and these matches are then aligned by a joint pose optimization module based on iteratively reweighted least squares. Temporal depth fusion is then performed given the estimated pose, but unfortunately, as one can see in this diagram, to exploit the benefits of depth fusion in this framework, CAPE has to run twice per frame, pushing even further the speed requirements of feature extraction. These novel modules for cylinder primitives are detailed below.

\subsection{Probabilistic Cylinder Fitting}

\begin{figure}
	\centering
	\includegraphics[scale=0.5]{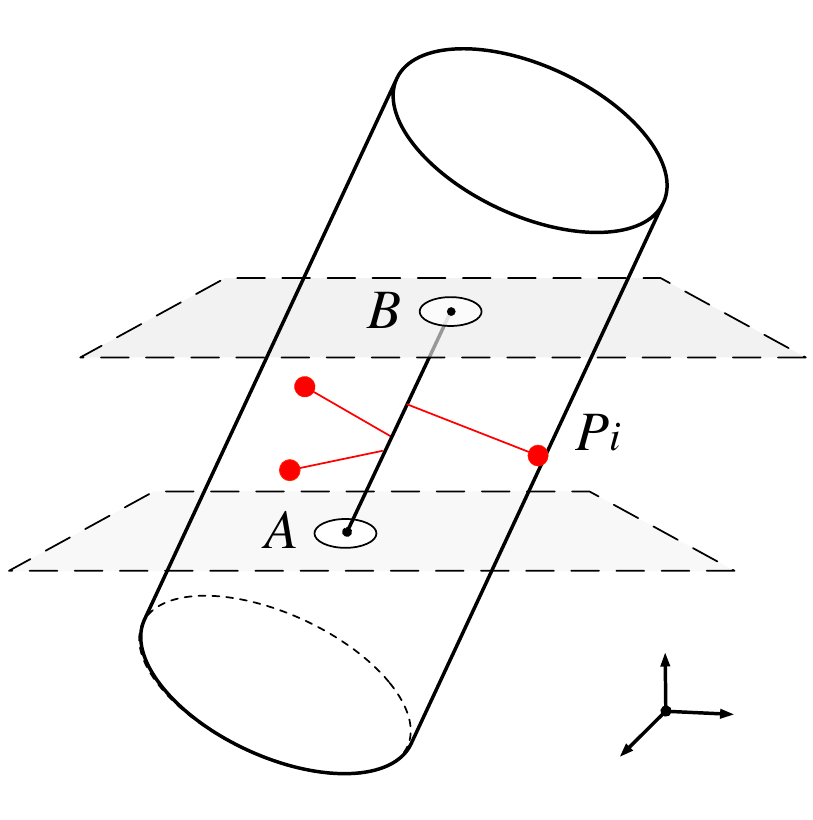}
	\caption{Iterative cylinder fitting solution. Planes represent the gauge constraint.}
	\label{fig6}
	\vspace*{-1mm} 
\end{figure}

To refine the parameters of the final cylinder segments by taking into account the segment pixels (instead of cells) and their depth uncertainties, we propose here an iterative probabilistic cylinder fitting based on non-linear weighted least squares, which is illustrated in Fig. \ref{fig6}. \par 
Here, a cylinder is represented by two points along the axis $\{A,B\}$ and the radius $r$. For estimating these, a minimal parameterization is possible by fixing one dimension for the two points. These are initialized using the center and the axis given by the solution in Section \ref{sec:plane_vs_cylinder}, and then we fix the 3D coordinate which has the largest range. As a result, the parameter vector has 5 dimensions in total, which are estimated by minimizing the sum of point to cylinder surface distances given by:
\begin{equation}
\label{eq:ils_cy}
E = \sum_{i}^{m} w_i\Big(\frac{\|(B - A)\times(A - P_i) \|}{\|B-A\|}   - r\Big)^2
\end{equation}

Ideally, the weights $w_i$ should be the inverse uncertainty of the residuals, however for simplicity and efficiency in this work we used the inverse of the depth uncertainties, as in \cite{proencca2017probabilistic}. The uncertainty of the parameters is then finally backpropagated using the Hessian approximation:
\begin{equation}
\label{eq:params_uncertainty}
\Sigma_{\xi} = (J_r^\top \Sigma_{r}^{-1} J_r)^{-1}
\end{equation}
where $J_r$ is the Jacobian matrix of the residuals wrt. the estimated parameters $\xi$, evaluated at the solution, and $\Sigma_{r}$ is a diagonal matrix containing the uncertainties of the residuals, which are found by first order propagation of the 3D point uncertainties through (\ref{eq:ils_cy}). It is worth noting that due to the coordinate constraint, the obtained uncertainties  for the two points are flat as shown in Fig. \ref{fig6}.
\par
In practice, the number of points per segment is too high, thus we subsample these using a grid with step size of 5 pixels. Although, significantly slower than the direct solution in Section \ref{sec:plane_vs_cylinder}, this solution takes around 4 iterations to converge using analytical Jacobians and a Levenberg-Marquardt solver, whereas with numerical Jacobian computation it takes around 50 iterations.
\subsection{Cylinder Matching}

For matching two cylinders between successive frames, we first check if the minimal angle formed by the two cylinder axis is less than a specified threshold (30$^\circ$) and if the Mahalanobis distance between the radii: $\frac{(r_1-r_2)^2}{\sigma^2_{r_1} +\sigma^2_{r_2}}$ is less than a maximum value, heuristically set to 2000. The radius uncertainties are extracted from (\ref{eq:params_uncertainty}). If a match passes these conditions, we then check if their image segment overlap is more than half of the size of the smallest segment, as proposed in \cite{proencca2017probabilistic} for plane segments.

\subsection{Pose Estimation based on Cylinders}
To estimate the relative camera pose: $\{R,t \mid R \in SO(3), t \in \mathbb{R}^3\}$, given a match between a cylinder with point parameters $\{A,B\}$ and a cylinder represented by $\{A',B'\}$, we express their error in the vector form as:
\begin{equation}
\label{eq:cylinder_r}
r_{c} = \begin{bmatrix} (B - A)\times(A - R A' -t) \\ (B - A)\times(A-  R B' - t)  \end{bmatrix}
\end{equation}
Effectively, this reflects the alignment between cylinder axes. For two plane matches with equations $\{N,d\}$ and $\{N',d'\}$, we make use of the plane-to-plane distance, described in \cite{proencca2017probabilistic}, such that, the residual can be derived, in the vector form, as: 
\begin{equation}
\label{eq:plane_r}
r_{p} = N'^\top R(N'^\top t+d') - dN^\top
\end{equation}
Let the set of plane matches be $\mathcal{P}$ and the set of cylinder matches be $\mathcal{C}$, then pose can be estimated by minimizing:
\begin{equation}
\label{eq:pose_optim}
\begin{aligned}
E = \alpha_{\textnormal{plane}}\sum_{p\in \mathcal{P}}r_p W_p r_p^\top + \alpha_{\textnormal{cylinder}}\sum_{c\in \mathcal{C}}r_{c}^\top W_{c}  r_{c}
\end{aligned}
\end{equation}
where $\alpha_{\textnormal{plane}}$ and $\alpha_{\textnormal{cylinder}}$ are two fixed scaling factors controlling the impact of the feature-types and $W_p$ and $W_c$ are diagonal weight matrices that are computed in every iteration as the inverse of the uncertainties of the residuals (\ref{eq:cylinder_r}) and (\ref{eq:plane_r}), which are obtained through first order error propagation of the feature parameter uncertainties, given by probabilistic model fitting, that is (\ref{eq:params_uncertainty}) for cylinders. For robustness, we combine (\ref{eq:pose_optim}) with reprojection errors of points and lines as in \cite{proencca2017probabilistic}.

\section{Experiments}

\begin{figure}[t]
 	\centering
 	\begin{tabular}{@{}c@{ }c@{ }c@{}}
 		\includegraphics[scale=0.12]{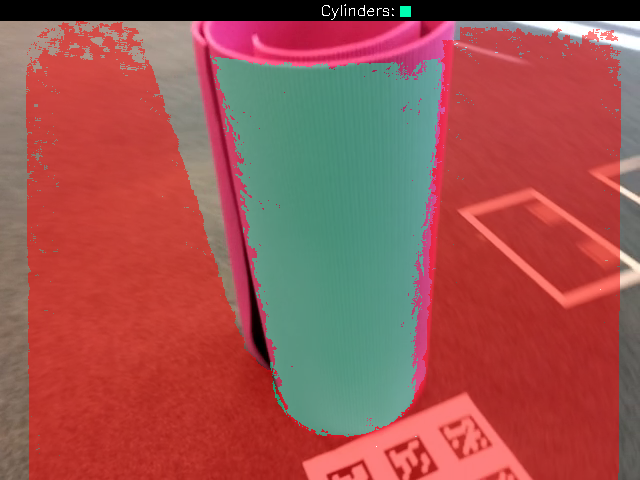} &
 		\includegraphics[scale=0.12]{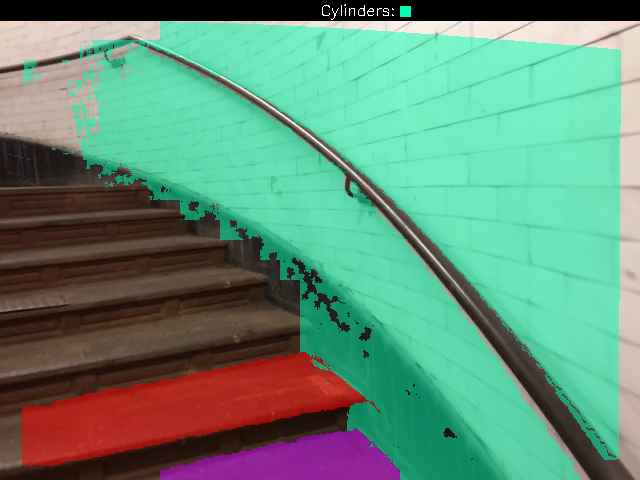} &
 		\includegraphics[scale=0.12]{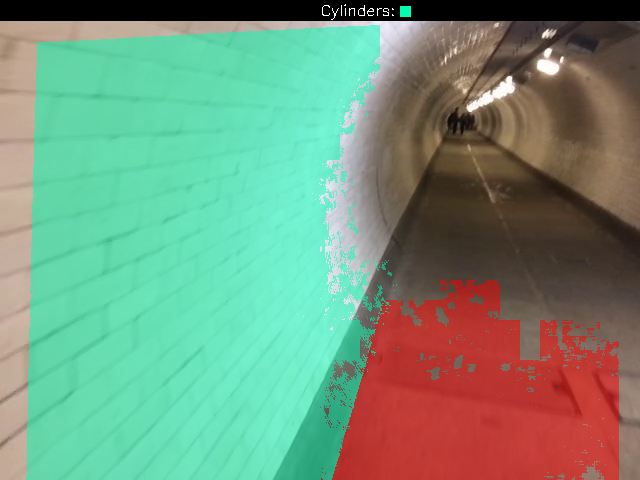}\\
 		yoga mat & spiral stairway & tunnel
 	\end{tabular} 
 	\caption{ Frames from the dataset collected in this work, processed by CAPE.}
 	\vspace*{-1mm} 
 	\label{fig7}
 \end{figure}

 We have conducted experiments on scenes with and without cylindrical surfaces. For non-cylindrical surfaces, we evaluate performance on a few sequences from the TUM RGB-D dataset \cite{tumdataset12iros} captured by structured-light Kinect and the synthetic ICL-NUIM dataset \cite{ICLNUIM}, whereas to capture cylindrical surfaces, we have collected 3 sequences with an Occipital Structure sensor, shown in Fig. \ref{fig7} and the supplementary video (see Fig. \ref{fig1}). A markerboard was used to measure the pose ground truth. While the shorter \textit{yoga\_mat} sequence contains ground truth for many frames, the other two only contain ground truth in the beginning and end of the trajectory as theses were captured in a close-loop. Timing results for the feature extraction are reported in the Section \ref{sec:timing} and the performance of VO is evaluated in Section \ref{sec:vo_performance}
 
 \subsection{Implementation Details}

All results were obtained on a PC with Intel i5-5257U CPU, using a single core. Depth maps were processed in VGA resolution.
Both CAPE and PEAC were used with a patch size of 20$\times$20 pixels. The remaining parameters for PEAC were left as default values, whereas CAPE parameters were sensibly set to: $T_N=cos(\pi/12)$, $plane\_min\_score=100$, $extrusion\_min\_score=100$, $k_1=5$, $k_2=5$ and the histogram has 400 bins.
In terms of VO, we used depth fusion with a maximum sliding window of 5 frames and combined feature points with the features given by CAPE or PEAC, except where noted. The scaling factors in (\ref{eq:pose_optim}) are fine-tuned in Section \ref{sec:vo_performance}.
 
  \subsection{Processing Time}
  \label{sec:timing}
\begin{figure}
	\centering
	\includegraphics[scale=0.43]{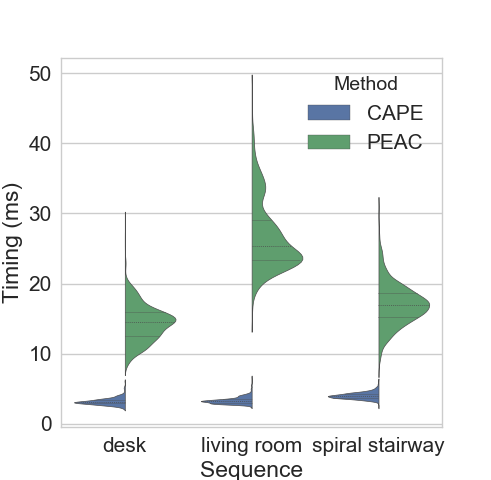}
	\caption{Timing results as violin plots for three sequences. The sequence \textit{desk} corresponds to the \textit{fr1\_desk} of \cite{tumdataset12iros} and the sequence \textit{living room} corresponds to the \textit{lr kt0} in \cite{ICLNUIM}.}
	\label{fig8}
	\vspace*{-1mm} 
\end{figure}

  The feature extraction timings are shown in Fig. \ref{fig8} for three sequences from three different datasets. CAPE performs consistently across the sequences taking on average 3 ms and the \textit{spiral\_stairway} sequence shows a negligible overhead introduced by cylinder segments. By contrast, PEAC, besides being 4-10 times slower on average, it exhibits a large variance and a heavy-tail. Interestingly, PEAC is significantly slower in the \textit{living room} sequence and one can also see a bimodal distribution. On close inspection, we observed that the large variances are due to the clustering and refinement steps and that clustering gets slower when the camera faces closely one wall, which increases the number of merging attempts by the clustering algorithm. Following this observation, we timed both CAPE and PEAC on a synthetic noise-free wall and obtain respectively: 2.6 and 34 ms with a 20$\times$20 patch and 5.8 and 300 ms with 10$\times$10 patch.
 
\subsection{Visual Odometry Performance}
 \label{sec:vo_performance}
\begin{figure}[t]
	\centering
		\includegraphics[scale=0.18]{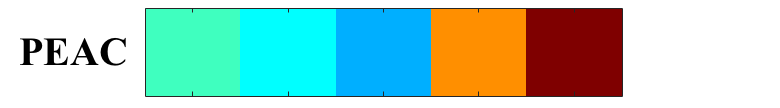}\\
		\includegraphics[scale=0.18]{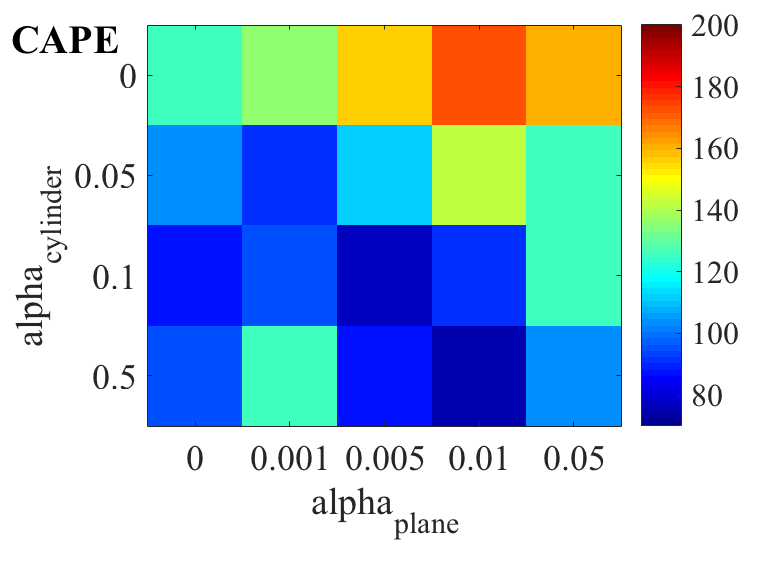}
	\caption{Impact of the joint pose estimation residual weights on the ATE  as RMSE in mm.}
	\vspace*{-1mm} 
	\label{fig9}
\end{figure}

\begin{figure}[t]
	\centering
	\begin{tabular}{@{}c@{\hspace{-0.5em}}c@{}}
	\includegraphics[scale=0.39]{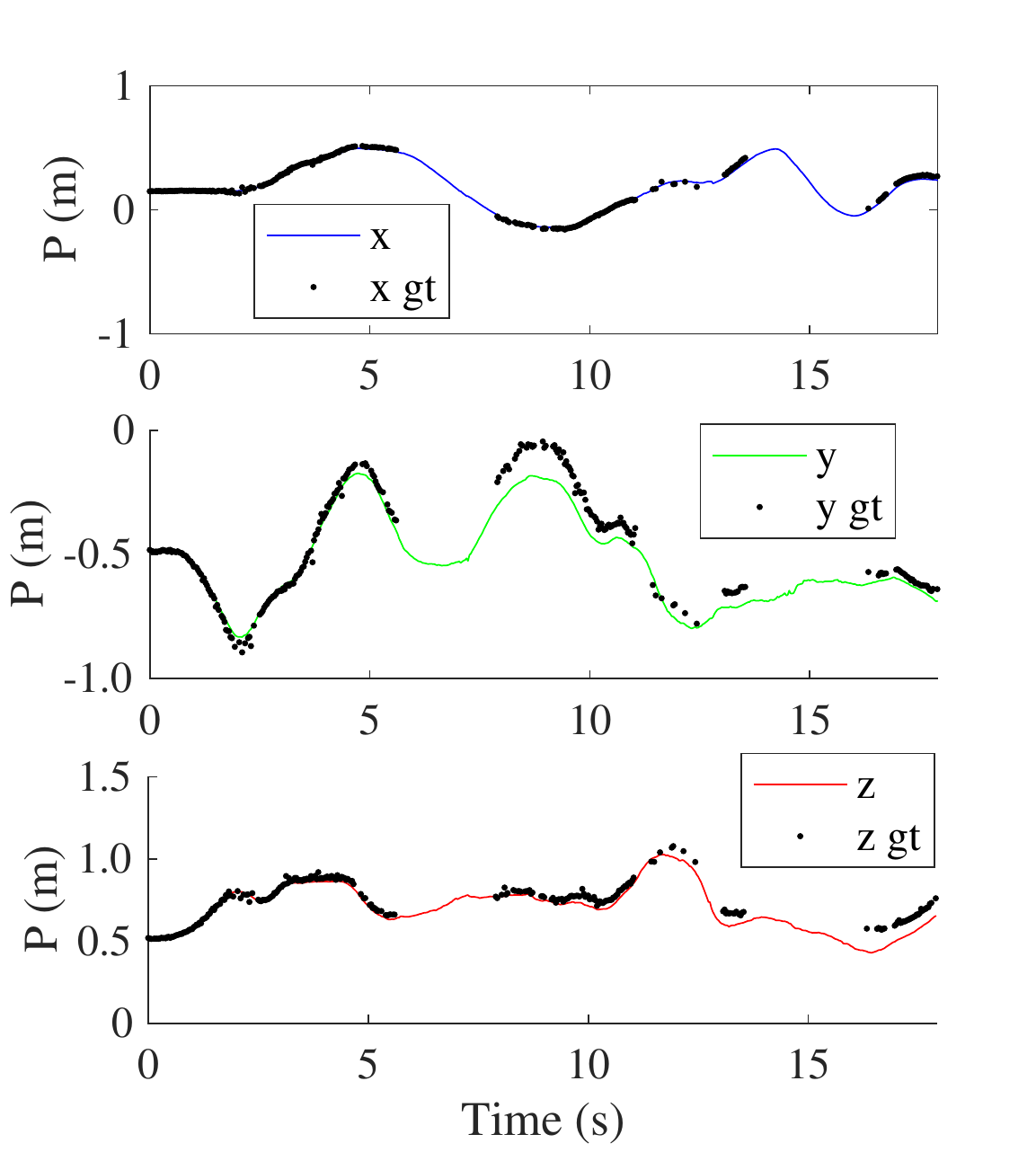}&
	\includegraphics[scale=0.39]{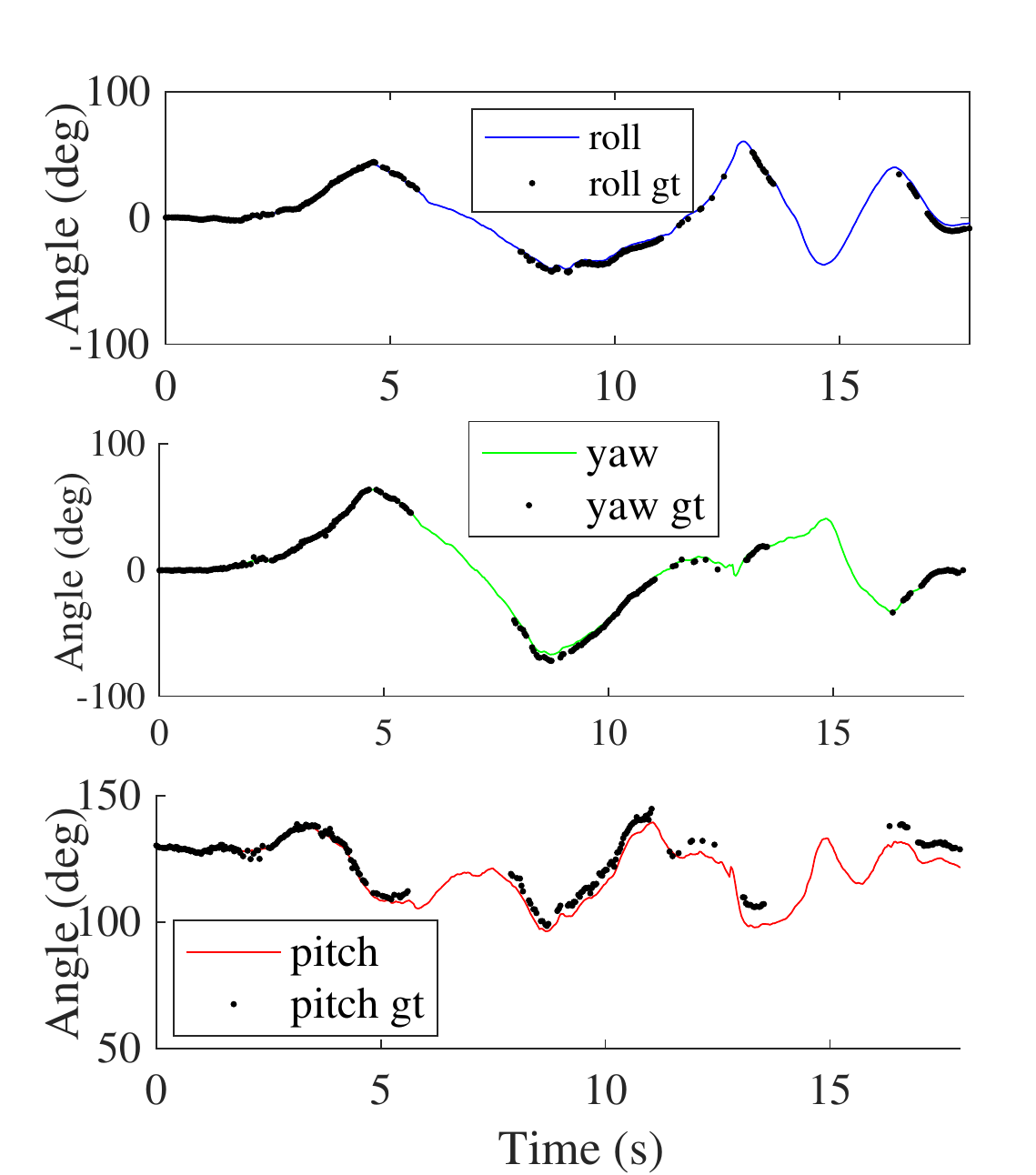}
\end{tabular}
	\caption{Ground truth trajectory of the \textit{yoga\_mat} sequence along with the path estimated by VO using CAPE.}
	\vspace*{-1mm} 
	\label{fig10}
\end{figure}

 First, the trajectory estimation error of VO is summarized in Fig. \ref{fig9} for the \textit{yoga mat} sequence while the impact of planes and cylinders on pose estimation, i.e., the feature-types weights in (\ref{eq:pose_optim}), are changed based on grid search. This is demonstrated as the absolute trajectory error (ATE) given the trajectory ground truth shown in Fig. \ref{fig10}. Setting all factors to zero means that the VO only uses point features. Up to $\alpha_{plane} = 0.05$, the performance of using planes extracted by PEAC is improved, but after that, the performance is severely degraded. By the contrary, CAPE with only planes fails to improve the performance as this discards the cylinder surface. This suggests that it is better to model this surface as a plane that not modelling at all. However, as we introduce cylinders extracted by CAPE, the error is consistently decreased for all plane weights, outperforming PEAC significantly. Based on Fig. \ref{fig9} and performance on other sequences, we fix $\alpha_{plane} = 0.01$ and $\alpha_{cylinder} = 0.1$ for the remaining experiments.
 
\begin{figure}
	\centering
	\begin{tabular}{@{\hspace{-0.3em}}c@{\hspace{-1em}}c@{}}
	\includegraphics[scale=0.32]{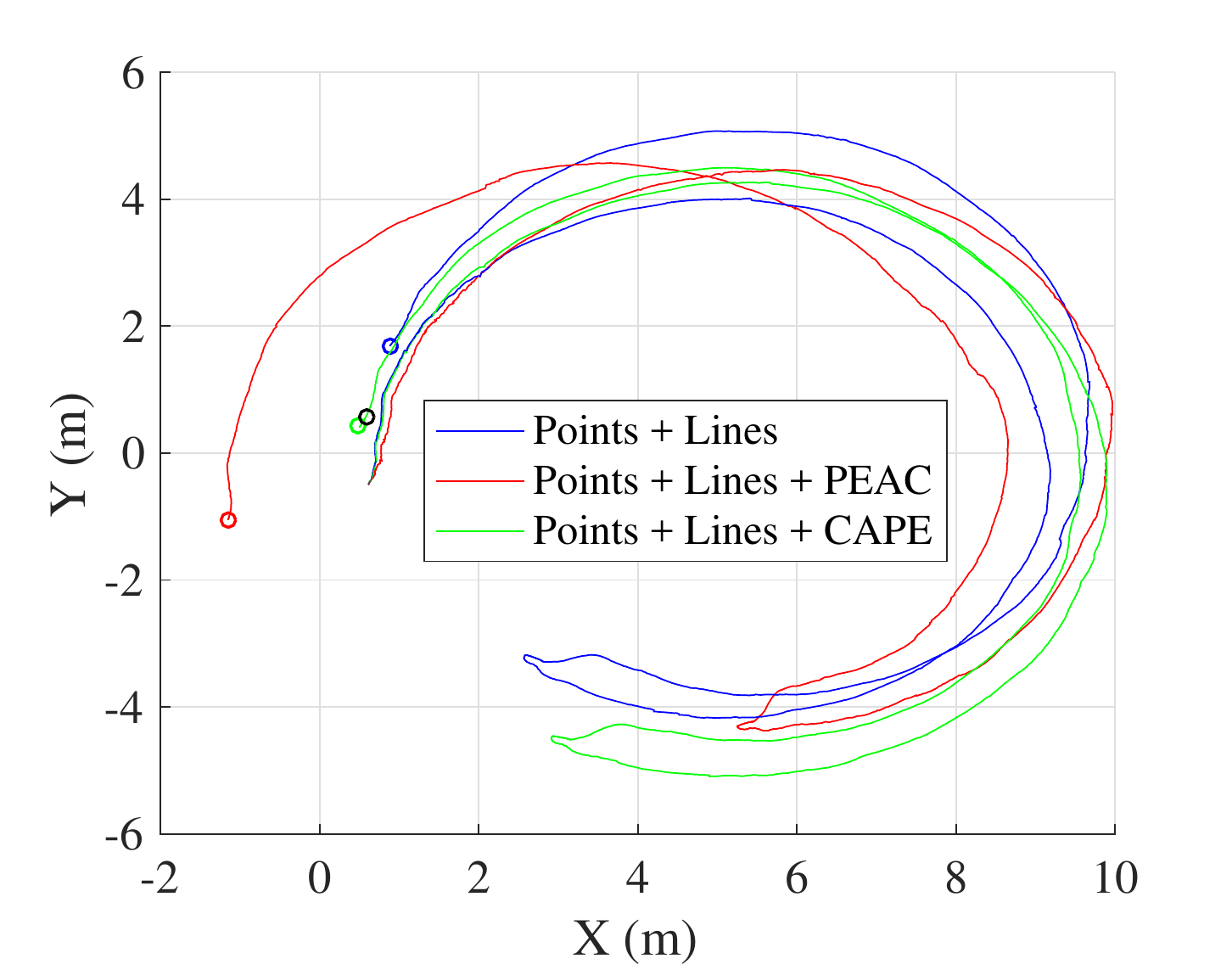} &
	\includegraphics[scale=0.32]{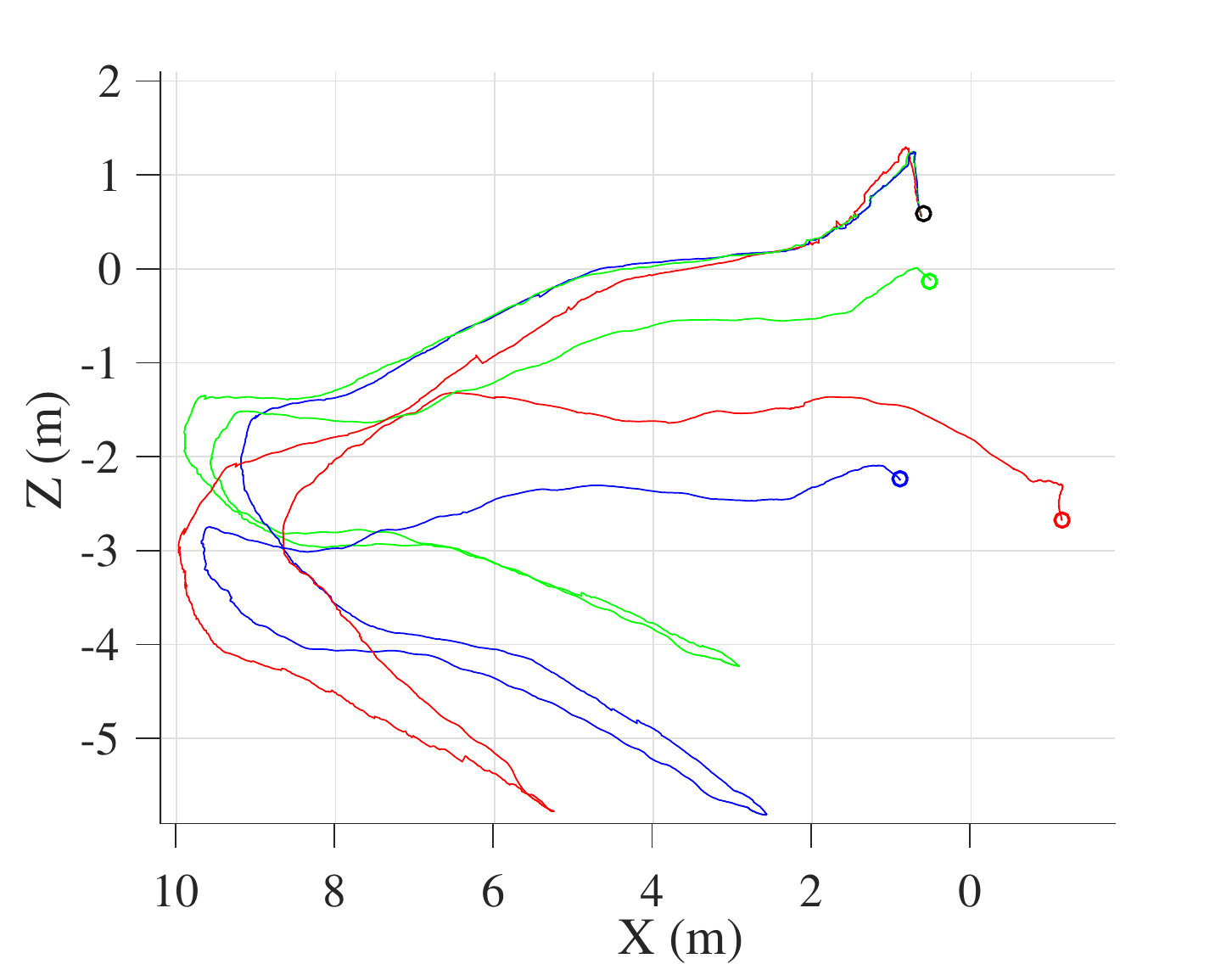}\\
	\includegraphics[scale=0.32]{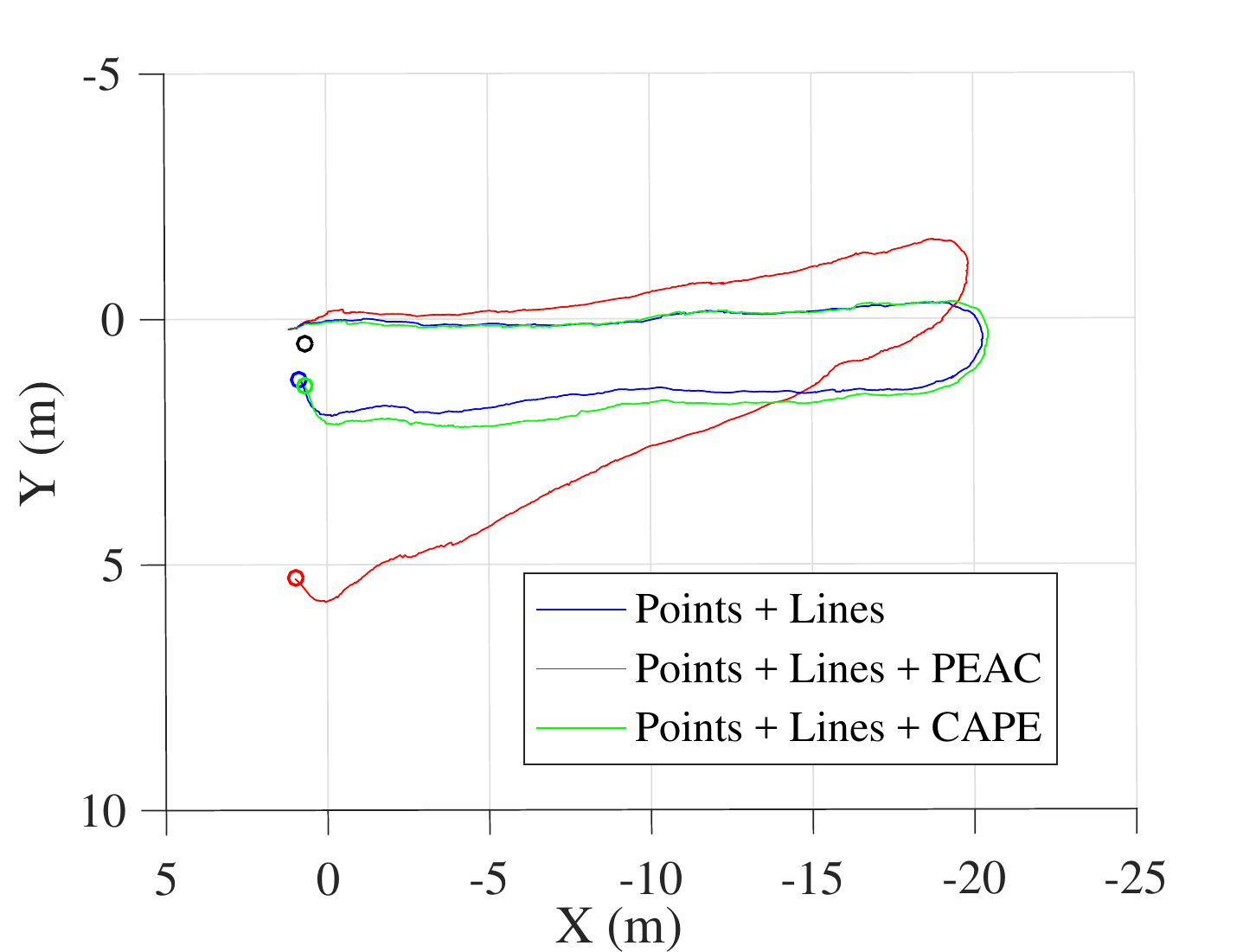} &
	\includegraphics[scale=0.32]{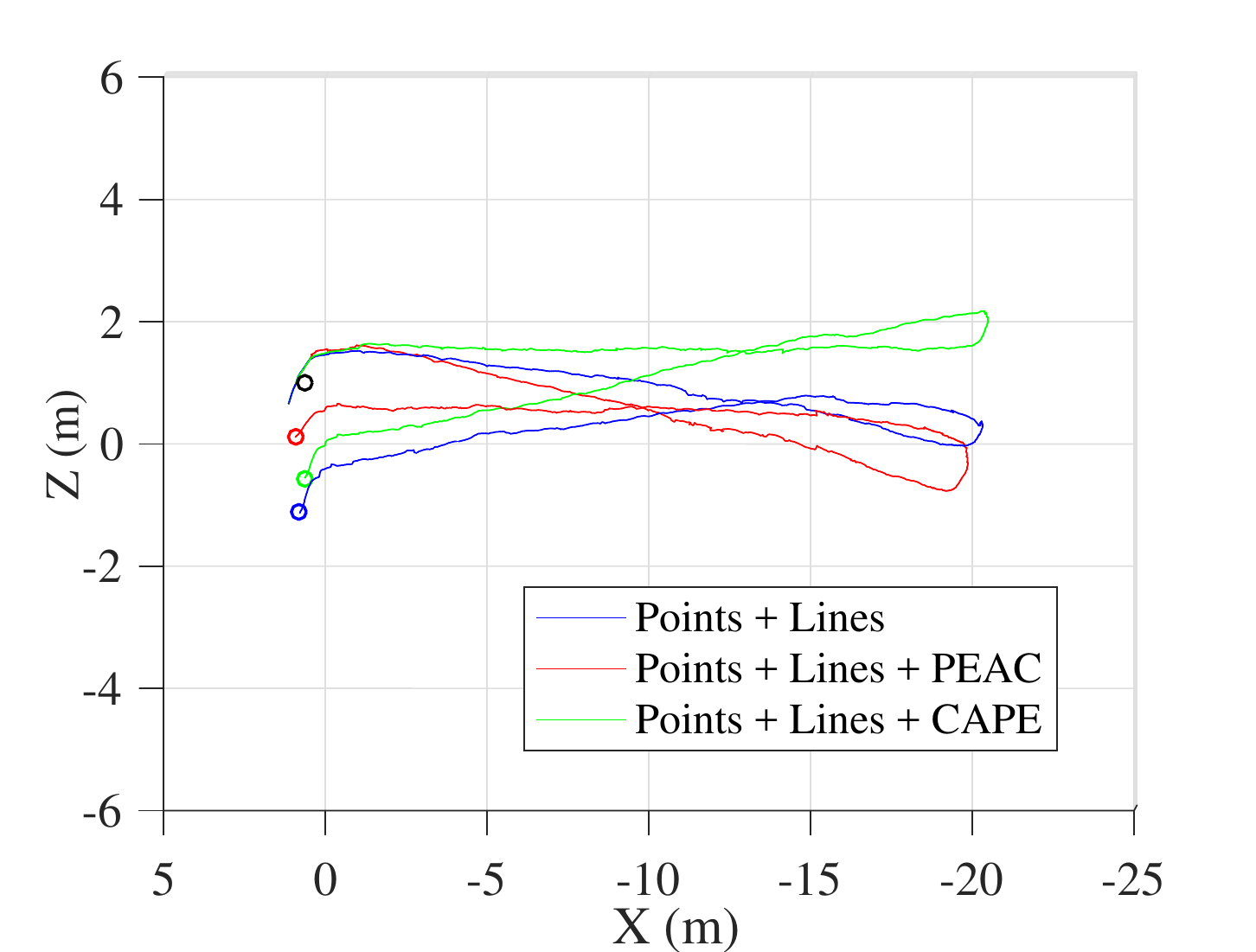}
	\end{tabular}
	\caption{Trajectories estimated on two sequences: Top row corresponds to the \textit{spiral\_stairway}. Bottom row corresponds to the \textit{tunnel}. Left column is the top view and right column is a side view. End of the trajectories are marked by a circle. The black one is the ground truth final position.}
	\label{fig11}
	\vspace*{-1mm} 
\end{figure}

The trajectories estimated for the other two sequences are shown in Fig. \ref{fig11} and the respective final errors are reported in Table \ref{tab:dataset_author}. We have found that combining just feature points with the planes and cylinders performed poorly on these environments as these are dominated by uniform surfaces, thus, here, we employed line segments as in the original system. While using PEAC, degrades the performance of the baseline, CAPE with cylinder primitives is able to improve the performance particularly on the \textit{spiral\_stairway}, where the user walked up and down the same stairs, thus trajectory should be two closely aligned lines in Fig. \ref{fig11}.

	\begin{table}[h]
		\centering
		\scriptsize{
			\begin{tabular}{|l|c|c|c|}
				\hline
				Seq. (distance) &  \begin{tabular}[c]{@{}c@{}}Points \& \\ Lines \end{tabular} &  \begin{tabular}[c]{@{}c@{}}Points \& \\ Lines \& PEAC \end{tabular} & \begin{tabular}[c]{@{}c@{}}Points \& \\ Lines \& CAPE \end{tabular} \\ \hline
				Tunnel (44 m) &  \begin{tabular}[c]{@{}c@{}}2.3 m \\ 23 deg\end{tabular} &  \begin{tabular}[c]{@{}c@{}}4.9 m \\  32 deg\end{tabular} & \begin{tabular}[c]{@{}c@{}}1.8 m \\  19 deg \end{tabular} \\ \hline
				Spiral stairway (51 m)  & \begin{tabular}[c]{@{}c@{}}3.0 m \\  6 deg\end{tabular} & \begin{tabular}[c]{@{}c@{}}4.2 m \\  25 deg\end{tabular} & \begin{tabular}[c]{@{}c@{}} 0.7 m \\ 6 deg\end{tabular} \\ \hline
		\end{tabular}}
		\caption{Final trajectory errors for the RGB-D sequences collected in this work.}
		\label{tab:dataset_author}
\end{table}

Results on cylinder-free scenes are reported in Table \ref{tab:dataset_public}. Although, performance is similar for several sequences, we can see in \textit{fr1\_360} that PEAC can outperform CAPE. We have noted that, with the specified parameters, PEAC can extract more planes far away from the camera than the selective CAPE, which can be advantageous for pose estimation when the number of features is critically low, as indicated by Fig. \ref{fig9}.

	\begin{table}[h]
	\centering
	\scriptsize{
		\begin{tabular}{|l|c|c|}
			\hline
			Seq. & Points \&  PEAC & Points \& CAPE \\ \hline
			fr1\_desk &  62 mm / 25 mm / 1.8 deg & 49 mm /  25 mm / 1.8 deg \\ \hline
			fr1\_360 & 117 mm / 68 mm / 3.3 deg & 131 mm /  73 mm / 3.3 deg \\ \hline
			fr3\_struct\_ntxt\_far & 80 mm / 36 mm / 0.9 deg & 80 mm / 35 mm / 0.9 deg \\ \hline
			or kt0 w/ noise & 209 mm / 7 mm / 0.5 deg & 160 mm / 7 mm / 0.5 deg \\ \hline
	\end{tabular}}
	\caption{Performance on non-cylindrical scenes from public datasets, in terms of RMSE, shown in the following order: ATE /  relative translational error / relative angular error of trajectory.}
	\label{tab:dataset_public}
\end{table}

\section{Conclusions and Limitations}

We demonstrated a consistently fast plane and cylinder extraction method that improves VO performance on scenes made of cylindrical surfaces. We believe this contribution can be further beneficial for full SLAM and model tracking systems. Operating on image cells instead of points, is key for efficiency and to deal with sensor noise, however this sacrifices resolution in the sense that surfaces smaller than the patch size are filtered out. Although this is not severe in VO as we are more concerned about large stable surfaces than smaller objects, which can be captured using feature points, this can be an issue for robotic grasping applications. Thus the parameters used here need further fine-tuning for other applications. Yet a more promising idea is to implement a multi-scale search. Another limitation, is that segments given by region growing that form more complex shapes (i.e. not extruded), are also removed. Therefore, incorporating other primitives such as spheres and cones remains as a worthwhile future direction. 


\bibliographystyle{ieeetr} 
{\footnotesize
\bibliography{iros_ref}

\begin{thebibliography}{10}

\bibitem{PlaneAndsurfels_2014}
R.~F. Salas-Moreno, B.~Glocken, P.~H. Kelly, and A.~J. Davison, ``Dense planar
  slam,'' in {\em IEEE International Symposium on Mixed and Augmented Reality
  (ISMAR)}, pp.~157--164, 2014.

\bibitem{Taguchi2013}
Y.~Taguchi, Y.-D. Jian, S.~Ramalingam, and C.~Feng, ``Point-plane slam for
  hand-held 3d sensors,'' in {\em IEEE International Conference on Robotics and
  Automation (ICRA)}, pp.~5182--5189, 2013.

\bibitem{kpaslam}
M.~Hsiao, E.~Westman, G.~Zhang, and M.~Kaess, ``Keyframe-based dense planar
  slam,'' in {\em IEEE International Conference on Robotics and Automation
  (ICRA)}, IEEE, 2017.

\bibitem{ma2016cpa}
L.~Ma, C.~Kerl, J.~St{\"u}ckler, and D.~Cremers, ``Cpa-slam: Consistent
  plane-model alignment for direct rgb-d slam,'' in {\em IEEE International
  Conference on Robotics and Automation (ICRA)}, pp.~1285--1291, IEEE, 2016.

\bibitem{proencca2017probabilistic}
P.~F. Proen{\c{c}}a and Y.~Gao, ``Probabilistic rgb-d odometry based on points,
  lines and planes under depth uncertainty,'' {\em Robotics and Autonomous
  Systems}.
\newblock In press, arXiv preprint:
  \href{https://arxiv.org/abs/1706.04034}{arXiv:1706.04034v3}.

\bibitem{feng2014fast}
C.~Feng, Y.~Taguchi, and V.~R. Kamat, ``Fast plane extraction in organized
  point clouds using agglomerative hierarchical clustering,'' in {\em IEEE
  International Conference on Robotics and Automation (ICRA)}, pp.~6218--6225,
  2014.

\bibitem{holz2013fast}
D.~Holz and S.~Behnke, ``Fast range image segmentation and smoothing using
  approximate surface reconstruction and region growing,'' in {\em Intelligent
  autonomous systems 12}, pp.~61--73, Springer, 2013.

\bibitem{trevor2013efficient}
A.~J. Trevor, S.~Gedikli, R.~B. Rusu, and H.~I. Christensen, ``Efficient
  organized point cloud segmentation with connected components,'' {\em Semantic
  Perception Mapping and Exploration (SPME)}, 2013.

\bibitem{yang2010plane}
M.~Y. Yang and W.~F{\"o}rstner, ``Plane detection in point cloud data,'' in
  {\em Proceedings of the 2nd int conf on machine control guidance, Bonn},
  vol.~1, pp.~95--104, 2010.

\bibitem{biswas2012depth}
J.~Biswas and M.~Veloso, ``Depth camera based indoor mobile robot localization
  and navigation,'' in {\em IEEE International Conference on Robotics and
  Automation (ICRA)}, pp.~1697--1702, 2012.

\bibitem{schnabel2007efficient}
R.~Schnabel, R.~Wahl, and R.~Klein, ``Efficient ransac for point-cloud shape
  detection,'' in {\em Computer graphics forum}, vol.~26, pp.~214--226, Wiley
  Online Library, 2007.

\bibitem{vosselman2004recognising}
G.~Vosselman, B.~G. Gorte, G.~Sithole, and T.~Rabbani, ``Recognising structure
  in laser scanner point clouds,'' {\em International archives of
  photogrammetry, remote sensing and spatial information sciences}, vol.~46,
  no.~8, pp.~33--38, 2004.

\bibitem{rabbani2005efficient}
T.~Rabbani and F.~Van Den~Heuvel, ``Efficient hough transform for automatic
  detection of cylinders in point clouds,'' {\em ISPRS WG III/3, III/4},
  vol.~3, pp.~60--65, 2005.

\bibitem{limberger2015real}
F.~A. Limberger and M.~M. Oliveira, ``Real-time detection of planar regions in
  unorganized point clouds,'' {\em Pattern Recognition}, vol.~48, no.~6,
  pp.~2043--2053, 2015.

\bibitem{poppinga2008fast}
J.~Poppinga, N.~Vaskevicius, A.~Birk, and K.~Pathak, ``Fast plane detection and
  polygonalization in noisy 3d range images,'' in {\em International Conference
  on Intelligent Robots and Systems (IROS)}, pp.~3378--3383, IEEE, 2008.

\bibitem{holz2011real}
D.~Holz, S.~Holzer, R.~B. Rusu, and S.~Behnke, ``Real-time plane segmentation
  using rgb-d cameras,'' in {\em Robot Soccer World Cup}, pp.~306--317,
  Springer, 2011.

\bibitem{salas2013slam++}
R.~F. Salas-Moreno, R.~A. Newcombe, H.~Strasdat, P.~H. Kelly, and A.~J.
  Davison, ``Slam++: Simultaneous localisation and mapping at the level of
  objects,'' in {\em Computer Vision and Pattern Recognition (CVPR), 2013 IEEE
  Conference on}, pp.~1352--1359, IEEE, 2013.

\bibitem{bowman2017probabilistic}
S.~L. Bowman, N.~Atanasov, K.~Daniilidis, and G.~J. Pappas, ``Probabilistic
  data association for semantic slam,'' in {\em Robotics and Automation (ICRA),
  2017 IEEE International Conference on}, pp.~1722--1729, IEEE, 2017.

\bibitem{hansen2013pipe}
P.~Hansen, H.~Alismail, P.~Rander, and B.~Browning, ``Pipe mapping with
  monocular fisheye imagery,'' in {\em Intelligent Robots and Systems (IROS),
  2013 IEEE/RSJ International Conference on}, pp.~5180--5185, IEEE, 2013.

\bibitem{zuliani2005multiransac}
M.~Zuliani, C.~S. Kenney, and B.~Manjunath, ``The multiransac algorithm and its
  application to detect planar homographies,'' in {\em Image Processing, 2005.
  ICIP 2005. IEEE International Conference on}, vol.~3, pp.~III--153, IEEE,
  2005.

\bibitem{torr2000mlesac}
P.~H. Torr and A.~Zisserman, ``Mlesac: A new robust estimator with application
  to estimating image geometry,'' {\em Computer vision and image
  understanding}, vol.~78, no.~1, pp.~138--156, 2000.

\bibitem{ICLNUIM}
A.~Handa, T.~Whelan, J.~McDonald, and A.~Davison, ``A benchmark for {RGB-D}
  visual odometry, {3D} reconstruction and {SLAM},'' in {\em IEEE International
  Conference on Robotics and Automation (ICRA)}, IEEE, 2014.

\bibitem{tumdataset12iros}
J.~Sturm, N.~Engelhard, F.~Endres, W.~Burgard, and D.~Cremers, ``A benchmark
  for the evaluation of rgb-d slam systems,'' in {\em International Conference
  on Intelligent Robots and Systems (IROS)}, IEEE, 2012.

\end{thebibliography}
}

\end{document}